\documentclass[10pt,twocolumn,letterpaper]{article}

\usepackage{cvpr}      

\usepackage{graphicx}
\usepackage{amsmath}
\usepackage{amssymb}
\usepackage{booktabs}
\usepackage{tabularx}
\usepackage{amsfonts}
\usepackage{amssymb}
\usepackage{caption}
\usepackage{capt-of,etoolbox}
\usepackage{scalerel}
\usepackage{makecell}
\usepackage{multirow}
\usepackage{dsfont}
\usepackage{pifont}
\usepackage{csquotes}
\usepackage[table]{xcolor}
\usepackage{xcolor}
\usepackage[normalem]{ulem}
\usepackage{appendix}

\usepackage[pagebackref,breaklinks,colorlinks]{hyperref}

\usepackage[capitalize]{cleveref}
\crefname{section}{Sec.}{Secs.}
\Crefname{section}{Section}{Sections}
\Crefname{table}{Table}{Tables}
\crefname{table}{Tab.}{Tabs.}
\def\SPSB#1#2{\rlap{\textsuperscript{#1}}\textsubscript{#2}}

\newcommand{\cmark}{\ding{51}}%
\newcommand{\xmark}{\ding{55}}%
\newcommand{\rom}[1]{\uppercase\expandafter{\romannumeral #1\relax}}
\definecolor{LightCyan}{rgb}{0.88,1,1}
\definecolor{LightYellow}{rgb}{1,1,0.88}
\definecolor{LightRed}{rgb}{1,0.88,1}

\def\cvprPaperID{15} 
\def\confName{CVPR}
\def\confYear{2022}

\begin{document}

\title{Towards Open-Set Object Detection and Discovery}

\author{Jiyang Zheng$^{\star \dagger}$ \qquad Weihao Li$^{\dagger}$ \qquad Jie Hong$^{\star \dagger}$ \qquad Lars Petersson$^{\dagger}$ \qquad Nick Barnes$^{\star}$ \\
$^{\star}$The Australian National University, Canberra, Australia \\ $^{\dagger}$Data61-CSIRO, Canberra, Australia\\
{\tt\small firstname.lastname@\{$^{\star}$anu.edu.au, $^{\dagger}$data61.csiro.au\}}
}
\maketitle

\begin{abstract}
With the human pursuit of knowledge, open-set object detection (OSOD) has been designed to identify unknown objects in a dynamic world. However, an issue with the current setting is that all the predicted unknown objects share the same category as \enquote{unknown}, which require incremental learning via a human-in-the-loop approach to label novel classes. In order to address this problem, we present a new task, namely Open-Set Object Detection and Discovery (OSODD). This new task aims to extend the ability of open-set object detectors to further discover the categories of unknown objects based on their visual appearance without human effort. We propose a two-stage method that first uses an open-set object detector to predict both known and unknown objects. Then, we study the representation of predicted objects in an unsupervised manner and discover new categories from the set of unknown objects. With this method, a detector is able to detect objects belonging to known classes and define novel categories for objects of unknown classes with minimal supervision. We show the performance of our model on the MS-COCO dataset under a thorough evaluation protocol. We hope that our work will promote further research towards a more robust real-world detection system.
\end{abstract}

\section{Introduction}
\label{sec:intro}
\begin{figure}[t]
  \centering
   \includegraphics[width=0.95\linewidth,page=10]{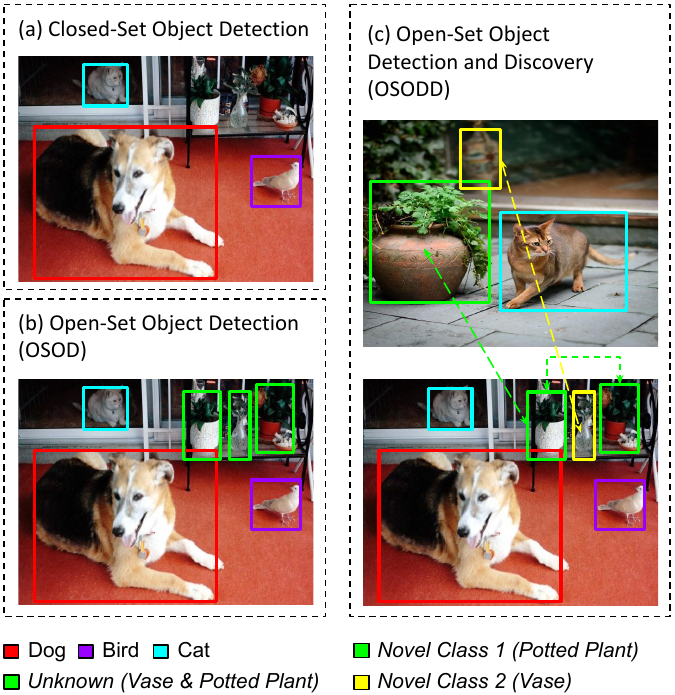}
    \caption{A visual comparison of object detection tasks. In closed-set detection, objects from unseen classes are ignored or incorrectly classified into the set of known classes. While in open-set object detection, unknown objects are localised but share the same category. Our task aims to detect objects of known classes and discover novel visual categories for the identified objects of unknown classes, which provides better scene understanding and a scalable learning paradigm.}
   \label{fig:fig_1}
\end{figure}

Object detection is the task of localising and classifying objects in an image. In recent years, deep learning approaches have advanced the detection models~\cite{girshick2015fast,redmon2016you,ren2016faster,he2017mask,cai2018cascade,wang2020frustratingly,carion2020end} and achieved remarkable progress. However, these methods work under a strong assumption that all object classes are known at the training phase. As a result of this assumption, object detectors would incorrectly treat objects of unknown classes as background or classify them as belonging to the set of known classes~\cite{dhamija2020overlooked} (see \cref{fig:fig_1}(a)). 

To relax the above closed-set condition, open-set object detection (OSOD)~\cite{miller2018dropout,dhamija2020overlooked,joseph2021towards} considers a realistic scenario where test images might contain novel classes that did not appear during training. OSOD aims at jointly detecting objects from the set of known classes and localising objects that belong to an unknown class. Although OSOD has improved the practicality of object detection by enabling detection of instances of unknown classes, there is still the issue that all identified objects of an unknown class share the same category as \enquote{unknown} (see \cref{fig:fig_1}(b)). Additional human annotation is required to incrementally learn novel object categories~\cite{joseph2021towards}.

Consider a child who is visiting a zoo for the first time. The child can recognise some animals that are seen and learned before, for example, `rabbit' or `bird', while the child might not recognise the species of many other rarely seen animals, like `zebra' and `giraffe'. After observing, the child's perception system will learn from these previously unseen animals' appearances and cluster them into different categories even without being told what species they are. 

In this work, we consider a new task, where we aim to localise objects of both known and unknown classes, assign pre-defined category labels for known objects, and discover new categories for objects of unknown classes (see \cref{fig:fig_1}(c)). We term this task \textit{Open-Set Object Detection and Discovery} (OSODD). We motivate our proposed task, OSODD, by suggesting that it is better suited to extracting information from images. New category discovery provides additional knowledge of data belonging to classes not seen before, helping intelligent vision-based systems to handle more realistic use cases. 


We propose a two-stage framework to tackle the problem of OSODD. First, we leverage the ability of an open set object detector to detect objects of known classes and identify objects of unknown classes. The predicted proposals of objects of known and unknown classes are saved to a memory buffer; Second, we explore the recurring pattern of all objects and discover new categories from objects of unknown classes. Specifically, we develop a self-supervised contrastive learning approach with domain-agnostic data augmentation and semi-supervised k-means clustering for category discovery.

\vspace{1mm}
Our contributions:
\begin{itemize}
    \item We formalise the task Open-Set Object Detection and Discovery (OSODD), which enables a richer understanding within real-world detection systems.
    \item We propose a two-stage framework to tackle this problem, and we present a comprehensive protocol to evaluate the object detection and category discovery performance.
    \item We propose a category discovery method in our framework using domain-agnostic augmentation, contrastive learning and semi-supervised clustering. The novel method outperforms other baseline methods in experiments.
\end{itemize}

\section{Related Work}
\label{sec:related}

\noindent \textbf{Open-Set Recognition.} Compared with closed-set learning, which assumes that only previously known classes are present during testing, open-set learning assumes the co-existence of known and unknown classes. Scheirer~\etal~\cite{scheirer2012toward} first introduce the problem of open-set recognition with incomplete knowledge at training time, \ie, unknown classes can appear during testing. They developed a classifier in a one-vs-rest setting, which enables the rejection of unknown samples. \cite{jain2014multi,scheirer2014probability} extend the framework in \cite{scheirer2012toward} to a multi-class classifier using probabilistic models with the extreme value theory to minimise fading confidence of the classifier. Recently, Liu~\etal~\cite{liu2019large} proposed a deep metric learning method to identify unseen classes for imbalanced datasets. Self-supervised learning~\cite{perera2020generative, gidaris2018unsupervised, tack2020csi} approaches have been explored to minimise external supervision.

\begin{table}
\centering
  \resizebox{1. \linewidth}{!}{
  \begin{tabular}{lccc}
    \toprule
    \textbf{Task} & \textbf{Dataset} &\textbf{Known classes} & \textbf{Unknown classes} \\ \midrule
    ODL & Open-Set & Non-Action      & Loc/Cat \\
    OSOD   & Open-Set & Detect    & Loc \\
    \midrule
    \textbf{OSODD (Ours)}& Open-Set & Detect & Loc/Cat \\
    \bottomrule
  \end{tabular}
  }
  \caption{Comparisons of different Object Detection and Discovery tasks. OSOD: open-set object detection; ODL: Object discovery and localization. \textit{Loc} means localise the objects of interest; \textit{Cat} means discover novel categories.} 
  \label{tab:task}
\end{table}

Miller~\etal~\cite{miller2018dropout} first investigate the utility of label uncertainty in object detection under open-set conditions using dropout sampling. Dhamija~\etal~\cite{dhamija2020overlooked} define the problem of open-set object detection~(OSOD) and conducted a study on traditional object detectors for their abilities in avoiding classifying objects of unknown classes into one of the known classes. An evaluation metric is also provided to assess the performance of the object detector under the open-set condition. \\

\noindent \textbf{Open-World Recognition.} 
The open-world setting introduced a continual learning paradigm that extends the open-set condition by assuming new semantic classes are introduced gradually at each incremental time step. Bendale~\etal~\cite{bendale2015towards} first formalise the open-world setting for image recognition and propose an open-set classifier using the nearest non-outlier algorithm. The model evolves when new labels for the unknown are provided by re-calibrating the class probabilities. 

Joseph~\etal~\cite{joseph2021towards} transfer the open-world setting to an object detection system and propose the task of open-world object detection~(OWOD). The model uses example replay to make the open-set detector learn new classes incrementally without forgetting the previous ones. The OWOD or OSOD model cannot explore the semantics of the identified unknown objects, and extra human annotation is required to learn novel classes incrementally. In contrast, our OSODD model can discover novel category labels for objects of unknown classes without human effort. \\

\noindent \textbf{Novel Category Discovery.} The novel category discovery task aims to identify similar recurring patterns in the unlabelled dataset. In image recognition, it was earlier viewed as an unsupervised clustering problem. Xie~\etal~\cite{xie2016unsupervised} proposed the deep embedding network that can cluster data and at the same time learn a data representation. Han~\etal~\cite{Han_2019_ICCV} formulated the task of novel class discovery~(NCD), which clusters the unlabelled images into novel categories using deep transfer clustering. The NCD setting assumes that the training set contains both labelled and unlabelled data, the knowledge learned on labelled data could be transferred to targeted unlabelled data for category discovery~\cite{zhong2021openmix,zhao2021novel, jia2021joint,han2021autonovel,fini2021unified}.

Object discovery and localisation ~(ODL)~\cite{lee2010object,cho2015unsupervised,li2018deep,rambhatla2021pursuit,Li_2019_CVPR_Workshops,chen2021channel} aims to jointly discover and localise dominant objects from an image collection with multiple object classes in an unsupervised manner. Lee and Grauman~\cite{lee2010object} used object-graph and appearance features for unsupervised discovery. Rambhat~\etal\cite{rambhatla2021pursuit} assumed partial knowledge of class labels and conducted the discovery leveraging a dual memory module. Compared to ODL, our OSODD both performs detection on previously known classes and discovers novel categories for unknown objects, which provide a comprehensive scene understanding.

Please refer to~\cref{tab:task} to see the summarised differences between our setting and other similar settings in the object detection problem.


\section{Task Format}
\label{sec:taskFormulation}
In this section, we formulate the task of Open-Set Detection and Discovery~(OSODD). We have a set of known object classes $C_k = \{C_1, C_2, \cdots, C_m \}$, and there exists a set of unknown visual categories $C_u = \{C_{m+1}, C_{m+2}, \cdots, C_{m+n} \}$, where $C_k \cap C_u = \varnothing$. The training dataset contains objects from $C_k$, and the testing dataset contains objects from $C_k \cup C_u$. An object instance $I$ is represented by $I = [c,x,y,w,h]$, denoting the class label ($c \in C_k$ or $C_u$), the top-left x, y coordinates, and the width and height from the centre of the object bounding box respectively. A model is trained to localise all objects of interest. Then, it classifies objects of a known class as one of $C_k^t$ and clusters objects of an unknown class into novel visual categories $C_u^t$.
 
\begin{figure*}
\begin{minipage}[b]{1.0\linewidth}
  \centerline{\includegraphics[width=\textwidth,page=3]{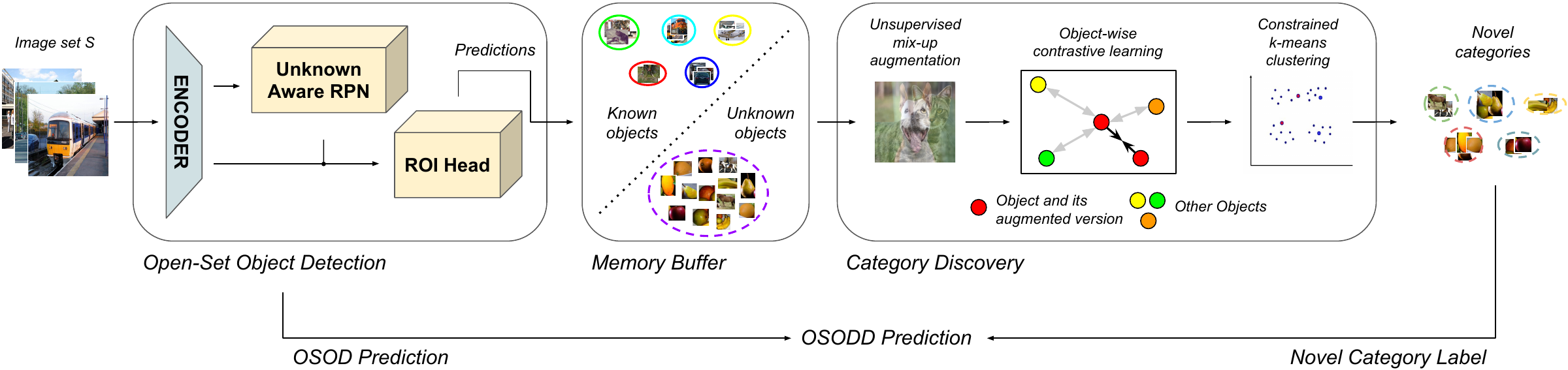}}
  \caption{Illustration of the two-stage method for Open-Set Object Detection and Discovery (OSODD). The first stage includes detecting objects of known classes and identifying objects of unknown classes using an open-set object detector. The instances of unknown classes are saved into the \textit{working memory} for category discovery. The instances of known classes are saved into the \textit{known memory} with their predicted semantic categories to assist the representation learning and clustering. The second stage pre-processes the objects from the memory buffer in an unsupervised manner. The representations of these saved objects are first learned in the latent space by contrastive learning, followed by a constrained k-means clustering used to find the novel categories beyond the known classes. Lastly, we update the open-set detection predictions with the novel category labels to generate the final OSODD prediction~(See visualisations in \cref{fig_3,fig_4}).}
  \label{fig_2}
\end{minipage}
\end{figure*}

\section{Our Approach}
This section describes our approach for tackling OSODD, beginning with an overview of our framework. We propose a generic framework consisting of two main modules, \textit{Object Detection and Retrieval (ODR)} and \textit{Object Category Discovery (OCD)} (see \cref{fig_2}).

The ODR module uses an open-set object detector with a dual memory buffer for object instances detection and retrieval. The detector predicts objects of known classes with their semantic labels from $C_k$ and the location information, where the unknown objects are localised but with no semantic information available. We store the predicted objects in the memory buffer~\cite{rambhatla2021pursuit}, which is used to explore novel categories. The buffer is divided into two parts: \textit{known memory} and \textit{working memory}. The known memory contains predicted objects of known classes with semantic labels; the working memory stores all current identified objects of unknown classes without categorical information. The model studies the recurring pattern of the objects from the memory buffer and discovers novel categories in the working memory. We assign the predicted objects of unknown classes from the detector with novel category labels using the discovered categories. The visualisation is shown in ~\cref{fig_4}.  

The OCD module explores the \textit{working memory} to discover new visual categories. It consists of an encoder component as the feature extractor and a discriminator which clusters the object representations. To train the encoder, we first retrieve the predicted objects from known classes saved in the known memory and the identified objects of unknown classes saved in working memory. Then, these instance samples are transformed using class-agnostic augmentation to create a generalised view over the data~\cite{cubuk2018autoaugment,zhong2020random,lee2020mix}. We use unsupervised contrastive learning where the predicted labels for the objects of known classes are ignored, the pairwise contrastive loss~\cite{oord2018representation} penalises dissimilarity of the same object in different views regardless of the semantic information. The contrastive learning enables the encoder to learn a more discriminating feature representation in the latent space~\cite{he2020momentum,chen2020simple}. Lastly, with the learned feature space from the encoder, the discriminator clusters the object embedding into novel categories by using the constrained k-means clustering algorithm~\cite{vaze2022generalized}. 


\subsection{Object Detection and Retrieval}
\label{sec:ODR}
\begin{figure}
  \centering
  \includegraphics[width=0.95\linewidth,page=2]{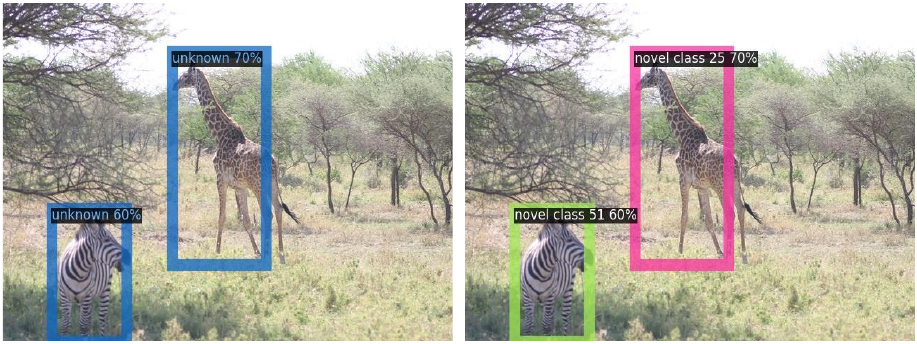}
    \caption{Comparison between OSOD and OSODD prediction. OSODD~(Right figure) has extended the OSOD~(Left figure) prediction by assigning novel category labels to instances of an unknown class.}
  \label{fig_3}
\end{figure}

\noindent\textbf{Open-Set Object Detector.}
An open-set object detector predicts the location of all objects of interest. Then it classifies the objects into semantic classes and identifies the unseen objects as unknown~(See `OSOD' in \cref{fig_3}). 

We use the Faster RCNN architecture~\cite{ren2016faster} as the baseline model, following ORE~\cite{joseph2021towards}. Leveraging the class-agnostic property of the region proposal network, we utilise an unknown-aware RPN to identify unknown objects. The unknown-aware RPN labels the proposals that have high scores but do not overlap with any ground-truth bounding box as the potential unknown objects. To learn a more discriminative representation for each class, we use a prototype based constrictive loss on the feature vectors $f_c$ generated by an intermediate layer in the ROI pooling head. A class prototype $p_i$ is computed by the moving average of the class instance representations, and the features $f_c$ of objects will keep approaching their class prototype in the latent space. The objective is formulated as:
\begin{equation}
\begin{aligned}
\ell_{pcl}(f_c)&= \sum_{i=0}^c \ell(f_c,p_i)\\
\ell(f_c,p_i) & = 
\begin{cases}
      \| f_c,p_i\| &\text{if } i=c \\
      \max~(0, \Delta - \| f_c,p_i\|) &\text{otherwise} 
\end{cases} 
\end{aligned}\label{eq:consloss}
\end{equation}
where $f_c$ is the feature vector of class $c$, $p_i$ is the prototype of class $i$, $\|f,p\|$ measures the distance between feature vectors and $\Delta$ is a fixed value that defines the maximum distance for dissimilar pairs. The total loss for the region of interest pooling is defined as: 
\begin{equation}
\begin{aligned}
\ell_{roi} &= \alpha_{pcl} \cdot \ell_{pcl} + \alpha_{cls} \cdot \ell_{cls} + \alpha_{reg} \cdot \ell_{reg}
\end{aligned}\label{eq:totalloss}
\end{equation}
where $\alpha_{pcl}$, $\alpha_{cls}$ and $\alpha_{reg}$ are positive adjustment ratios. $\ell_{cls}$, $\ell_{reg}$ are the regular classification and regression loss.

Given the encoded feature $f_c$, we use an open-set classifier with an energy-based model~\cite{lecun2006tutorial} to distinguish the objects of known and unknown classes. The trained model is able to assign low energy values to known data and thus creates dissimilar representations of distribution for the objects of known and unknown classes. When new known class annotations are made available, we utilise the example replay to alleviate forgetting the previous classes.\\

\noindent\textbf{Memory Module.}
As described above, we propose to use a dual memory module to store predicted instances for category discovery. The open-set detector detects the objects of interest with their locations and the predicted label. The objects of a known class $I_{k}$ are saved into known memory $M_k$ with their semantic labels $c \in C_k$. These objects are treated as a labelled dataset for the following category discovery. The identified objects of an unknown class $I_{u}$ are stored in the working memory $M_w$. We perform the category discovery on $M_w$, which aims to assign all instances in $M_w$ with a novel category label $c \in C_u$. We update the open-set object detector's prediction using the novel category labels and produce our final OSODD predictions.

\subsection{Object Category Discovery}
\label{sec:OCD}
\noindent\textbf{Category Number Estimation.}
Our category discovery approach requires an estimation of the number of potential classes. We use the class estimation method from~\cite{Han_2019_ICCV}, one of the most commonly used techniques for image-level novel category discovery. The model uses a k-means clustering method to estimate the category number in the target dataset without any parametric learning. The generalisation ability of the method towards our problem has been evaluated in~\cref{sec:estimateResult}.\\

\noindent\textbf{Representation Learning.}
Representation learning aims to learn more discriminative features for input samples. We adapt contrastive learning~\cite{oord2018representation} and utilise objects from both known and working memory to help the network to learn an informative embedding space. The learning is conducted in an unsupervised manner. Following~\cite{he2020momentum}, we build a dynamic dictionary to store samples. The network is trained to maximise similarities for positive pairs (an object and its augmented version) while minimising similarity for negative pairs (different object instances) in the embedding space. For an object representation, the contrastive loss is formulated as~\cite{chen2020improved}:
\begin{equation}
\begin{aligned}
\ell_{q,\{k\}} &= -\log \frac{\exp (q \cdot k^+/\tau)}{\exp (q \cdotp k^+/\tau) + \sum\limits_{k^-} \exp (q \cdotp k^-/\tau)}
\end{aligned}\label{eq:constraiveloss}
\end{equation}
where $q$ is a query object representation, \{k\} is the queue of key object samples, $k^+$ is an augmented version of $q$, known as the positive key, and $k^-$ is the representations of other samples, known as the negative key. $\tau$ is a temperature parameter. On top of the contrastive learning head, we adopt an unsupervised augmentation strategy~\cite{lee2020mix} which replaces all samples with mixed samples. It minimises the vicinal risk~\cite{chapelle2000vicinal} which discriminates classes with very different pattern distributions and create more training samples~\cite{zhang2017mixup}. For each sample in the queue $\{k\}$, we combine it with the query object representation $q$ via linear interpolation and generate a new view $k_{m,i}$. Correspondingly, a new virtual label $v_i$ for the $i$th mix sample $x_{m,i}$ is defined as:
\begin{equation}
    v_i= 
\begin{cases}
    1,& \text{if } q \text{ and } k^+ \text{ are chosen;}\\
    0, & \text{otherwise;}
\end{cases}\label{eq:label}
\end{equation}
where $q$ and $k^+$ are the positive sample pairs, the virtual label is assigned to $1$ if the mixing pair are from the same object instance.\\

\noindent\textbf{Novel Category Labelling.}
Using the encoded representation of the objects, we perform the label assignments using constrained k-means clustering~\cite{vaze2022generalized}, a non-parametric semi-supervised clustering method. The constrained k-means clustering takes object encoding from both known and working memory as its input. It converts the standard k-means clustering into a constraint algorithm by forcing the labelled object representation to be hard-assigned to their ground-truth class. In particular, we treat the object instances from the known memory $M_k$ as the labelled samples. We manually calculate the centroid for each labelled class. These centroids from $M_k$ serve as the first group of initial centroids for the k-means algorithm. We then randomly initialise the rest of the centroids for novel categories using the k-mean++ algorithm~\cite{arthur2006k}. For each iteration, the labelled object instances are assigned to the pre-defined clusters while the unknown object instances from $M_w$ are assigned to the cluster with the minimal distance between the cluster centroids and the object embedding. By doing this, we effectively avoid falsely predicted objects (\ie objects that belong to one of the semantic classes being predicted as unknown) from influencing the centroid update. We run the last cluster assignment step using only the novel centroids to ensure that all unknown objects from working memory are assigned to a discovered visual category in the final prediction. The novel centroids from the algorithm represent the discovered novel categories.

\section{Experimental Setup}
\label{sec:experimentssetup}
We provide a comprehensive evaluation protocol for studying the performance of our model in detecting objects from known classes and discovery of new novel categories for objects of unknown classes in our target dataset.
\subsection{Benchmark Dataset}
\label{sec:benchmark}
Pascal VOC 2007~\cite{everingham2010pascal} contains 10k images with 20 labelled classes. MS-COCO~\cite{lin2014microsoft} contains around 80k training and 5k validation images with 80 labelled classes. These two object detection datasets are used to build our benchmark. Following the setting of open-world object detection~\cite{zhao2022revisiting}, the classes are separated into known and unknown for three tasks  $\mathcal{T} = \{T_1,T_2,T_3\}$. For task $T_t \in \mathcal{T}$, all known classes from $\{T_i \text{ }|\text{ } i<t\}$ are treated as known classes for $T_t$ while the remaining classes are treated as unknown. For the first task $T_1$, we consider 20 VOC classes as known classes, and the remaining non-overlapping 60 classes in MS-COCO are treated as the unknown classes. New classes are added to the known set in the successive tasks,\ie, $T_2$ and $T_3$. For evaluation, we use the validation set from MS-COCO except for 48 images that are incompletely labelled~\cite{zhao2022revisiting}. We summarise the benchmark details in~\cref{tab:example}.

\begin{table}
  \centering
  \resizebox{0.88 \linewidth}{!}{
  \begin{tabular}{c|ccc}
    \toprule
    &Task-1&Task-2&Task-3\\
    \midrule
    \makecell{Semantic Split}&\makecell{VOC Classes}&\makecell{Outdoor,Accessory,\\Appliance, Truck\\ Wild Animal}& \makecell{Sports,\\Food}\\
    \midrule
    \makecell{Known/Unknown Class}&\makecell{20/60}&\makecell{40/40}&\makecell{60/20}\\
    \midrule
    \makecell{Training Set}&\makecell{16551} &\makecell{45520} &\makecell{39402} \\
    \midrule
    \makecell{Validation Set}&\multicolumn{3}{c}{\makecell{1000}}\\
    \midrule
    \makecell{Test Set}&\multicolumn{3}{c}{\makecell{4952}}\\
    \bottomrule
  \end{tabular}}
  \caption{Details of class split for the Benchmark. Task-1, Task-2 and Task-3 have different dataset splits of known and unknown classes.}
  \label{tab:example}
\end{table}

\subsection{Evaluation Metrics.}
\label{sec:eval}
\noindent\textbf{Object Detection Metrics.} 
A qualified open-set object detector needs to accurately distinguish unknown objects~\cite{dhamija2020overlooked}. UDR~(Unknown Detection Recall)~\cite{zhao2022revisiting} is defined as the localisation rate of unknown objects, and UDR~(Unknown Detection Precision)~\cite{zhao2022revisiting} is defined as the rate of correct rejection of objects of an unknown class. Let true-positives~(TP\textsubscript{u}) be the predicted unknown object proposals that have intersection over union IoU $>0.5$ with ground truth unknown objects. Half false-negatives~(FN\SPSB{*}{u}) be the predicted known object proposals that have IoU $>0.5$ with ground truth unknown objects. False-negatives(FN\textsubscript{u}) is the missed ground truth unknown objects. UDR and UDP are calculated as follow:
\begin{equation}
\begin{aligned}
\text{UDR} &= \frac{\text{TP\textsubscript{u}}+\text{FN\SPSB{*}{u}}}{\text{TP\textsubscript{u}}+\text{FN\textsubscript{u}}}\\
\text{UDP} &= \frac{\text{TP\textsubscript{u}}}{\text{TP\textsubscript{u}}+\text{FN\SPSB{*}{u}}}\\
\end{aligned}\label{eq:udrdup}
\end{equation}
In our task, the other important aspect is to localise and classify objects of interest from the known classes. We evaluate the closed-set detection performance using the standard mean average precision~(mAP) at IoU threshold of 0.5~\cite{ren2016faster}. To show the incremental learning ability, we provide the mAP measurement for the newly introduced known classes and previously known classes separately~\cite{peng2020faster,joseph2021towards}.\\

\noindent\textbf{Category Discovery Metrics.} 
Category discovery can be evaluated using clustering metrics~\cite{Han_2019_ICCV,zhong2021neighborhood,rambhatla2021pursuit,vaze2022generalized,lee2010object,hong2022goss}. We adopt the three most commonly used clustering metrics for our object-based category discovery performance. Suppose a predicted proposal of an object of an unknown class has matched to a ground truth unknown object. Let the predicted category label of the object proposal be $\hat{y_i}$, the ground truth label for the object is denoted as $y_i$. We calculate the clustering accuracy~(ACC)~\cite{Han_2019_ICCV} by:
\begin{equation}
\begin{aligned}
\text{ACC} &= \max_{p \in P_y} \frac{1}{N} \sum_{i=1}^N \mathds{1} \{y_i = p(\hat{y_i})\}
\end{aligned}\label{eq:ACC}
\end{equation}
where $N$ is the number of clusters, and $P_y$ is the set of all permutations of the unknown class labels.

Mutual Information $I(X,Y)$ quantifies the correlation between two random variables $X$ and $Y$. The range of $I(X,Y)$ is from $0$~(Independent) to $+\infty$. Normalised mutual information~(NMI)~\cite{strehl2002cluster} is bounded in the range $[0,1]$. Let $Cl$ be the set of ground truth classes, and $\widehat{Cl}$ be the set of predicted clusters. The NMI is formulated as: 
\begin{equation}
\begin{aligned}
\text{NMI} &= \frac{I(Cl, \widehat{Cl})}{[H(Cl)+H(\widehat{Cl})]/2}
\end{aligned}\label{eq:NMI}
\end{equation}
where $I(Cl, \widehat{Cl})$ is the sum of mutual information between each class-cluster pair. $H(Cl)$ and $H(\widehat{Cl})$ compute the entropy using maximum likelihood estimation. The Purity of the clusters is defined as:
\begin{equation}
\begin{aligned}
\text{Purity} &= \frac{1}{N} \sum_{i=1}^N \max\limits_k |Cl_k \cap \widehat{Cl_i} |
\end{aligned}\label{eq:Purity}
\end{equation}
Here, $N$ is the number of clusters and $\max$ is the highest count of objects for a single class within each cluster.

\section{Results and Analysis}
\label{sec:result}
\subsection{Baselines}
\label{sec:baseline}
\noindent\textbf{Object Detection Baselines.} 
Our framework uses an open-set object detector for known and unknown instance detection. We compare two recent approaches: Faster-RCNN+ ~\cite{joseph2021towards} and ORE~\cite{joseph2021towards}. The Faster-RCNN+ is a popular two-stage object detection method, which is modified from Faster RCNN\cite{ren2016faster} to localise objects of unknown classes by additionally adapting an unknown-aware regional proposal. ORE uses contrastive clustering and an energy-based classifier to discriminate the representations of known and unknown data. Our generic framework could cooperate with any open-set object detector, hence it is highly flexible. \\

\noindent\textbf{Category Discovery Baselines.} 
We compare our novel method with three baseline methods, including k-means, FINCH~\cite{sarfraz2019efficient} and a modified approach from DTC~\cite{Han_2019_ICCV}. 

\textit{K-means} clustering is a non-parametric clustering method that minimises within-cluster variances. In every iteration, the algorithm first assigns the data points to the cluster with the minimum pairwise squared deviations between samples and centroids; then, it updates cluster centroids with the current data points belonging to the cluster.

\textit{FINCH}~\cite{sarfraz2019efficient} is a parameter-free clustering method that discovers linking chains in the data by using the first nearest neighbour. The method directly develops the grouping of data without any external parameters. To make a fair comparison, we set the number of clusters to the same as the other baseline methods. We discuss the performance of FINCH in estimating the number of novel classes in \cref{sec:estimateResult}.

\textit{DTC+}, the DTC method~\cite{Han_2019_ICCV} is proposed for NCD problems~\cite{han2020automatically}, where the setting assumes the availability of unlabelled data at the training phase. The algorithm modifies deep embedded clustering~\cite{xie2016unsupervised} to learn knowledge of the labelled subset and transfer it to the unlabelled subset. This setting requires the unlabelled data in the training and testing set to be from the same classes. However, no unknown instances are available in training under the open-set detection setting. Hence, the NCD-based approaches, such as DTC cannot be directly applied to our problem. To facilitate the method in our settings, we modify it by transferring a portion of the classes from the known memory to the working memory during training and treating them as additional unknown classes. We evaluate DTC's generalisation performance on our problem in \cref{sec:discoveryresult}.

\subsection{Experimental Results}
We report the quantitative results of the novel category number estimation, object detection and novel category discovery performance in~\cref{sec:estimateResult,sec:OSODresult,sec:discoveryresult}. We show and discuss the qualitative results in~\cref{fig_4} and in the supplementary material.
\subsubsection{Novel Category Number Estimation}
\label{sec:estimateResult}
We show the results of estimating the number of novel categories in~\cref{tab:CatEst}. The middle two columns show the automatically discovered grouping by the FINCH algorithm~\cite{sarfraz2019efficient}. The numbers are under-estimated by a large margin of 30\%, 32.5\% and 40\% respectively. The last two columns show the result using DTC~\cite{Han_2019_ICCV}. It is found that the estimated number was lower than the ground truth class number, with an average error rate of 21\%. By exploring the ground-truth labels in the grouping, we found that both methods tend to ignore object classes with a small number of samples. Compared to the class estimation in the image recognition task~\cite{Han_2019_ICCV,vaze2022generalized}, the detection task faces more biased datasets as well as fewer available samples. Hence, it is still a challenging task for object category estimation.

\begin{table}
\centering
  \resizebox{0.7 \linewidth}{!}{
  \begin{tabular}{cc|cc|cc}
    \toprule
    Task&GT&FINCH\cite{sarfraz2019efficient}&Error&Est.\cite{Han_2019_ICCV}&Error\\
    \midrule
    1&60&42&30\%&48&20\%\\
    2&40&27&32.5\%&31&22.5\%\\
    3&20&12&40\%&16&20\%\\
    \bottomrule
  \end{tabular}
  }
  \caption{Result of novel Category estimation.}
  \label{tab:CatEst}
\end{table}

\subsubsection{Open-Set Object Detection}
\label{sec:OSODresult}
We compare two baseline models for the object detection part in our framework and show the result in \cref{tab:OSOD}. For each task, we record the mAP of all objects to evaluate the closed-world detection result. UDR and UDP reflect the unknown objectness performance and discrimination performance. The ORE outperforms the modified Faster-RCNN on known classes detection by a smaller margin, which are $-0.14\%, +1.14\%$ and $+1.01\%$ respectively. The mAP scores get lower when new semantic classes are being introduced. The UDR result shows that ORE performs better on unknown object localisation, with a $+0.95\%$ average unknown detection rate. As opposed to closed-set detection, the UDR scores improved when more classes are made available to the model. The Faster-RCNN baseline can only localise objects of an unknown class, but it does not identify them from known classes hence there is no UDP score. 

\subsubsection{Novel Category Discovery}
\label{sec:discoveryresult}
Results of the object category discovery are shown in~\cref{tab:est} and~\cref{tab:gt}. The test condition is the same as the open-set detection. Our discovery method is able to accurately explore novel categories among the objects of unknown classes.

Using the estimated number of classes, the discovery results are reported in~\cref{tab:est}. We observe that our method outperformed other baseline methods in the first two tasks. In Task-3, where there are 60 known classes and 20 unknown classes, our accuracy and purity score is slightly lower than the FINCH algorithm by $0.8\%$ and $0.1\%$. We suggest that Task-3 may contain more biased unknown object classes, therefore becoming challenging for self-supervised learning to learn generalised representations. 

We report the results using the ground truth number of classes in~\cref{tab:gt}. The results shown are similar to ~\cref{tab:est}, where our method has the best-aggregated performance over three tasks. The method achieves respectable quantitative results considering the challenging level of the task. 

\begin{table}
\begin{minipage}{1.0\linewidth}
  \centering
 \resizebox{1. \linewidth}{!}{
  \begin{tabular}{lccccccccc}
    \toprule
    & \multicolumn{3}{c}{Task-1} & \multicolumn{3}{c}{Task-2} & \multicolumn{3}{c}{Task-3}\\
    \cmidrule(lr){2-4}\cmidrule(lr){5-7}\cmidrule(lr){8-10}
    Method &mAP& UDR & UDP &mAP & UDR & UDP  &mAP  & UDR & UDP\\
    \midrule
    F-RCNN +  & -/ \textbf{56.16} &\textbf{20.14}&- &51.09/ 23.84  &21.54&- &35.69/ 11.53  &30.01&-\\
    ORE \cite{joseph2021towards}  & -/ 56.02  &20.10 &\textbf{36.74} &\textbf{52.19}/ \textbf{25.03} &\textbf{22.63}&\textbf{21.51} &\textbf{37.23}/ \textbf{12.02} &\textbf{31.82}&\textbf{23.55}\\
    \bottomrule
  \end{tabular}
  }
  \caption{Baseline model comparison for open-set detectors. The  mean average precision (mAP) is recorded for the previous/current known objects, there is no previous known for Task-1.}
  \vspace{4mm}
  \label{tab:OSOD}
  \end{minipage}
\end{table}

\begin{table}
  \begin{minipage}{1.0\linewidth}
  \centering
  \resizebox{1. \linewidth}{!}{
  \begin{tabular}{c|ccc|ccc|ccc}
    \toprule
    & \multicolumn{3}{c|}{Task-1} & \multicolumn{3}{c|}{Task-2} & \multicolumn{3}{c}{Task-3}\\
    \cmidrule(lr){2-4}\cmidrule(lr){5-7}\cmidrule(lr){8-10}
    Method &NMI   &ACC  &Purity &NMI &ACC  &Purity  &NMI  &ACC  &Purity \\
    \midrule
    K-means  & \color{gray}{\textbf{8.5}}  &5.3 & \color{gray}{\textbf{9.3}}&\color{gray}{\textbf{5.0}}&6.2&\color{gray}{\textbf{12.0}} &\color{gray}{\textbf{5.3}}&10.9&27.6\\
    FINCH~\cite{sarfraz2019efficient}  &2.8  &\color{gray}{\textbf{6.0}} &8.2 &5.4&\color{gray}{\textbf{6.3}}&9.9&\color{gray}{\textbf{5.3}}&{\textbf{17.2}}& {\textbf{29.4}}\\
    DTC+~\cite{Han_2019_ICCV}   &7.5  &4.6 &5.2&4.0&4.2&7.5&3.9&5.0&25.4\\
    \midrule
    Ours & {\textbf{11.0}} & {\textbf{6.3}} & {\textbf{12.6}}& {\textbf{5.8}}& {\textbf{6.9}}& {\textbf{13.3}}& {\textbf{6.5}}&\color{gray}{\textbf{16.4}}&\color{gray}{\textbf{29.3}}\\
    \bottomrule
  \end{tabular}}
  \caption{Results of discovery with estimated class number (48, 31, 16 for Task-1, Task-2 and Task-3 respectively). The highest score in each column is bold in black, and the second-highest score in each column is bold in {\color{gray}{grey}}. Our novel method has outperformed the proposed baseline models for all scores in Task-1 and Task-2. The cluster accuracy and purity scores are the second-highest in Task-3, with a marginal difference to the best-performed baseline.}
  \vspace{4mm}
  \label{tab:est}
  \end{minipage}
\end{table}
\begin{table}
   \begin{minipage}{1.0\linewidth}
   \centering
  \resizebox{1. \linewidth}{!}{
  \begin{tabular}{c|ccc|ccc|ccc}
    \toprule
    & \multicolumn{3}{c|}{Task-1} & \multicolumn{3}{c|}{Task-2} & \multicolumn{3}{c}{Task-3}\\
    \cmidrule(lr){2-4}\cmidrule(lr){5-7}\cmidrule(lr){8-10}
    Method &NMI   &ACC  &Purity &NMI &ACC  &Purity  &NMI  &ACC  &Purity \\
    \midrule
    K-means  &\color{gray}\textbf{11.9}  &6.0 &12.4  &\color{gray}\textbf{5.9} &6.1 &12.8  &\color{gray}\textbf{6.0}&11.6&27.9\\
    FINCH~\cite{sarfraz2019efficient} &10.3  &\color{gray}\textbf{6.1} &\color{gray}\textbf{12.5} &4.8&\color{gray}\textbf{7.5}&\color{gray}\textbf{13.4} &5.5&\textbf{13.6}&\color{gray}\textbf{28.3}\\
    DTC+~\cite{Han_2019_ICCV}   &8.3  &4.7 &9.2&4.2&5.0&12.1&5.0&7.7&26.1\\
    \midrule
    Ours &\textbf{13.1} &\textbf{6.5} &\textbf{13.1}&\textbf{7.0}&\textbf{7.5}&\textbf{13.8} &\textbf{6.1}&\color{gray}\textbf{13.2}&\textbf{29.1}\\
    \bottomrule
  \end{tabular}}
  \caption{Results of discovery with ground truth class number (60, 40, 20 for Task-1, Task-2 and Task-3 respectively). The highest score in each column is bold in black, and the second-highest score in each column is bold in {\color{gray}{grey}}. With the pre-defined number of classes, our method has achieved the highest scores for all three tasks, except for the accuracy in Task-3, which is behind the highest scoring baseline method by a small margin. The overall performance of our method is the best among all the proposed baselines.}
  \label{tab:gt}
  \end{minipage}
\end{table}

\begin{figure*}
\centering
\begin{minipage}[b]{1\linewidth}
  \centering
  \centerline{\includegraphics[width=\textwidth,page=1]{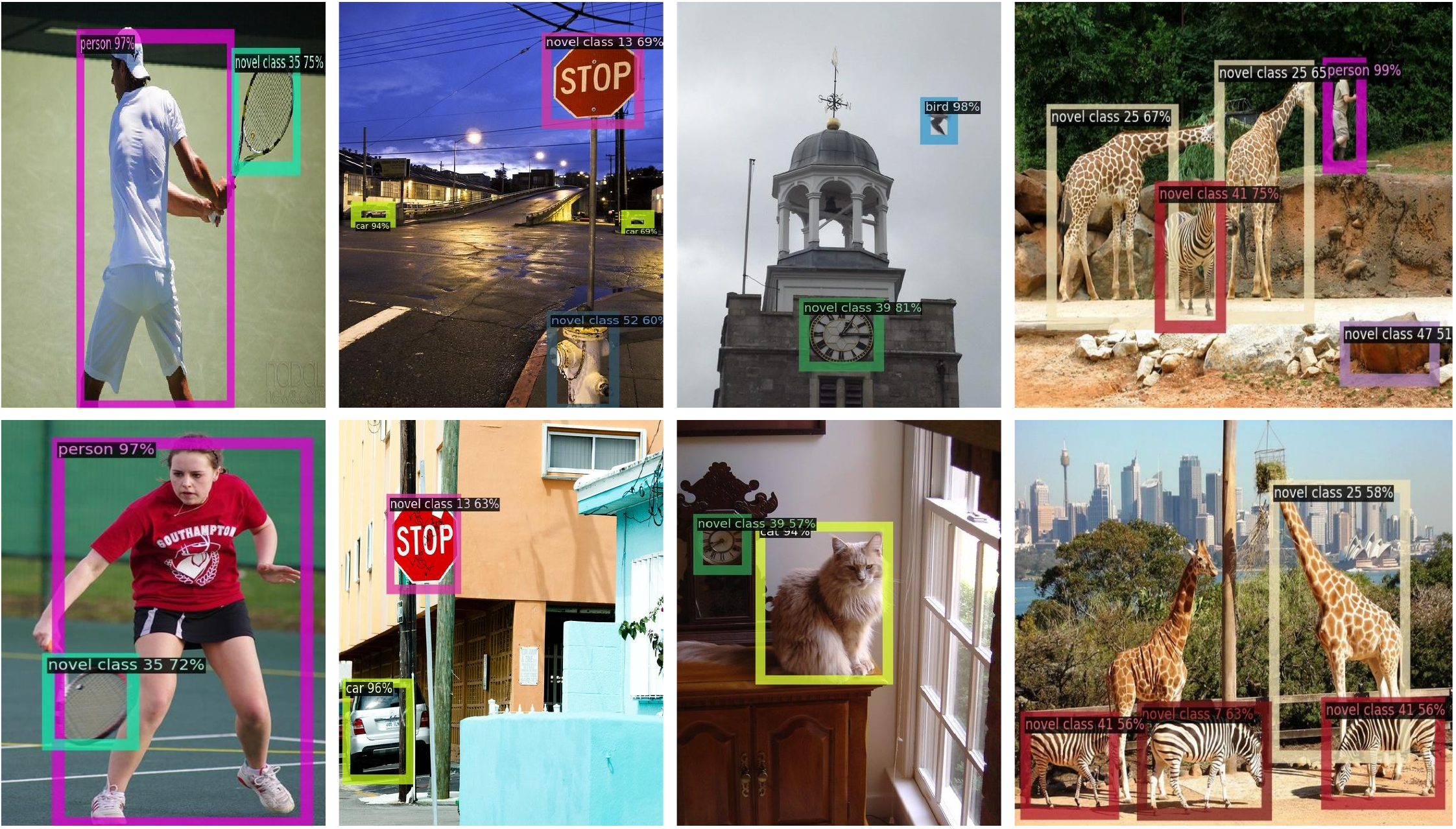}}
\end{minipage}
\caption{Visualisation of OSODD predictions for Task-1. The \textit{tennis racket}, \textit{stop sign}, \textit{fire hydrant}, \textit{clock}, \textit{giraffe} and \textit{zebra} are the novel classes that have not been introduced at this stage. The same bounding box colour indicates objects that belong to the same class or novel category. The last column demonstrates a failure case where a giraffe is not detected, and one of the zebras is assigned to the wrong visual category. More visualised results are provided in the supplementary material.}
\vspace{4mm}
  \label{fig_4}
\end{figure*}

\begin{table*}
\centering
  \resizebox{1 \linewidth}{!}{
  \begin{tabular}{c|ccc|ccc|ccc|ccc}
    \toprule
    &\multicolumn{2}{c}{Representation Learning}&Category Discovery & \multicolumn{3}{c|}{Task-1} & \multicolumn{3}{c|}{Task-2} & \multicolumn{3}{c}{Task-3}\\
    \cmidrule(lr){2-3}\cmidrule(lr){4-4}\cmidrule(lr){5-7}\cmidrule(lr){8-10}\cmidrule(lr){11-13}
    &Mix-Up Augmentation & Contrastive Learning &Semi-supervised Clustering & NMI   &ACC  &Purity &NMI &ACC &Purity  &NMI  &ACC  &Purity \\
    \midrule
    \rom{1}&\xmark&\xmark&\cmark &8.9  &5.6 &10.5  &4.7 &5.4 &11.9  &5.5 &14.7 &27.7\\
    \rom{2}&\xmark&\cmark&\cmark &10.5  &6.3 &12.0  &5.6 &5.4 &13.2  &6.1 &15.5 &28.6\\
    \rom{3}-1&\cmark&\cmark&\xmark &9.6  &5.7 &11.7  &5.2 &6.3 &12.9  &5.8 &15.9 &28.8\\
    \rom{3}-2&\cmark&\cmark&\xmark &7.4  &6.3 &12.3  &5.4 &6.4 &13.1  &6.0 &\textbf{16.8} &28.7\\
    \midrule
    \rom{4}&\cmark &\cmark &\cmark &\textbf{11.0} &\textbf{6.3} &\textbf{12.6}&\textbf{5.8}&\textbf{6.9}&\textbf{13.3}&\textbf{6.5}&16.4&\textbf{29.3} \\
    \bottomrule
  \end{tabular}
  }
  \caption{Ablation Study on components of our proposed category discovery method. The complete method with all the proposed modules achieves the best-aggregated performance in all tasks, which shows the importance of each component contributing to the method.}
  \label{tab:ablation1}
\end{table*}

\subsection{Ablation study}
\label{sec:ablation}
To study the contribution of each component in our proposed framework, we design ablation experiments and show the results in~\cref{tab:ablation1}.\\

\noindent\textbf{Representation Learning.} The effects of the representation learning in discovering novel classes are shown in Cases \rom{1}, \rom{2} and \rom{4}. The clustering result without encoding is reported in Case \rom{1}. The result with only contrastive learning is reported in Case \rom{2}. We observe that the performance without encoding is around 10\% lower compared to Case \rom{4} which is our method. Contrastive learning without the mix-up argumentation reflects higher scores compared to Case \rom{1}, but it is still around 4\% lower in the aggregated scores compared to Case \rom{4}. This suggests that representation learning is critical for constructing a strong baseline.\\

\noindent\textbf{Category Discovery.} We evaluate the effects of using semi-supervised clustering in Case \rom{3}-1, \rom{3}-2 and \rom{4}. In Case \rom{3}-1. We make the clustering algorithm fully unsupervised by removing the labelled centroids and instances. The results decrease by around 8\% in all tasks. Since the FINCH algorithm~\cite{sarfraz2019efficient} shows a competitive result in~\cref{tab:est} and~\cref{tab:gt}. In Case \rom{3-2}, we replace the semi-supervised clustering with the FINCH algorithm. The results show that Case \rom{4} outperforms Case \rom{3-2} in the task aggregation scores, which indicates our model better clusters the samples with the same learned feature space. \\

\noindent\textbf{Memory Module.} To show the effects of the current memory design, we ablate the module by removing the known memory in representation learning. We report the results in the supplementary material.

\section{Conclusion}
In this work, we propose a framework to detect known objects and discover novel visual categories for unknown objects. We term this task Open-Set Object Detection and Discovery~(OSODD), as a natural extension of open-set object detection tasks. We develop a two-stage framework and a novel method for label assignment, outperforming other popular baselines. Compared to detection and discovery tasks, OSODD can provide more comprehensive information for real-world practices. We hope our work will contribute to the object detection community and motivate further research in this area.
    
{\small
\bibliographystyle{ieee_fullname}
\bibliography{egbib}
}

\clearpage

\begin{appendices}
\begin{center}
      \Large\textbf{Appendix.}
\end{center}
\section{Implementation}
\label{sec:implementation}
For the object detection part, we rebuild ORE~\cite{joseph2021towards} with ResNet-50~\cite{he2016deep} as the backbone network. The initial learning rate is set to $0.01$. The weight decay is $1e^{-4}$. The momentum $\eta$, the margin parameter $\Delta$ and the temperature parameter $\mathcal{T}$ are empirically set to $0.9$, $15$ and $1$ respectively. The unknown object and non-maximum suppression threshold are set to $0.5$ and $0.4$. We train the model on three successive tasks (\ie, $T_1$, $T_2$, $T_3$ or Task-1, Task-2, Task-3), with 8 epochs on each task. The experiments are conducted on NVIDIA Tesla P100 $4$ GPUs with a batch size of $128$. For object category discovery, we select ResNet-50~\cite{he2016deep} as the backbone with no pre-trained weights. The network encodes the object instance into $256-$dimension using a linear projection head in the last layer. The learning rate is $0.015$~\cite{goyal2017accurate} with $200$ epochs for each training circle.

\subsection{Semantic Split}
The detailed known classes and unknown classes split of each task in MS-COCO~\cite{lin2014microsoft}, and Pascal VOC~\cite{everingham2010pascal} are shown in \cref{tab:t1}. The intersected classes between COCO and VOC are treated as the known classes for the first task, Task-1.

\section{Ablation study on Memory Module}
\label{sec:ablation2}
This section provides an ablation study on the memory module to show the effects of the known memory in representation learning. The results are reported in~\cref{tab:ablation2}. The performance of our method with only working memory is shown in Case \rom{1} where the detected known objects at the training phase will not be included in the representation learning. Compared to Case \rom{2} where our model is using both working and known memory, we can see that the performance of \rom{1} is worse over three tasks. It shows the design of our memory module is important for class discovery as it allows the model to learn more generalised embedding representations.

\section{Mutual Information and Entropy}
\label{sec:MI}
We have introduced the normalised mutual information for clustering performance evaluation. Here, we provide the formulation for the two major components in the normalised mutual information, which are the mutual information~(MI) and the entropy~(H). Let $Cl$ be the set of ground truth classes, and $\widehat{Cl}$ be the set of predicted clusters. The MI and entropy are formulated as~\cite{paninski2003estimation}:
\begin{equation}
\begin{aligned}
I(Cl, \widehat{Cl}) &=\sum\limits_k \sum\limits_j P(Cl_k \cap \widehat{Cl}_j) \log \frac{P(Cl_k \cap \widehat{Cl}_j)}{P(Cl_k) P(\widehat{Cl}_j)}\\
H(Cl) &= -\sum\limits_k P(Cl_k) \log P(Cl_k)\\
H(\widehat{Cl}) &= -\sum\limits_j P(\widehat{Cl}_j) \log P(\widehat{Cl}_j)
\end{aligned}\label{eq:MI}
\end{equation}
where $P(Cl_k)$, $P(\widehat{Cl}_j)$ and $P(Cl_k \cap \widehat{Cl}_j)$ are the probabilities of a object being in cluster $Cl_k$, $\widehat{Cl}_j$ and $Cl_k \cap \widehat{Cl}_j$ respectively. The probability is calculated as the number of corresponding objects divided by the total number of instances.

\begin{table}
\centering
  \resizebox{1. \linewidth}{!}{
  \begin{tabular}{c|cc|ccc|ccc|ccc}
    \toprule
    &\multicolumn{2}{c}{Memory Module} & \multicolumn{3}{c|}{Task-1} & \multicolumn{3}{c|}{Task-2} & \multicolumn{3}{c}{Task-3}\\
    \cmidrule(lr){2-3}\cmidrule(lr){4-6}\cmidrule(lr){7-9}\cmidrule(lr){10-12}
    &Known & Unknown & NMI   &ACC  &Purity &NMI &ACC &Purity  &NMI  &ACC  &Purity \\
    \midrule
    \rom{1}&\xmark&\cmark &8.8  &5.6 & 9.9&5.2&6.3&12.4 &5.5&11.7&28.2\\
    \midrule
    \rom{2}&\cmark &\cmark  &\textbf{11.0} &\textbf{6.3} &\textbf{12.6}&\textbf{5.8}&\textbf{6.9}&\textbf{13.3}&\textbf{6.5}&\textbf{16.4}&\textbf{29.3} \\
    \bottomrule
  \end{tabular}
  }
  \caption{Ablation Study on Memory Module.}
  \label{tab:ablation2}
\end{table}

\begin{table*}[!htp]
\centering
  \resizebox{1 \linewidth}{!}{
  \begin{tabular}{cccccccccc}
    \toprule
    \rowcolor{LightRed}
    \multicolumn{10}{c}{Task-1}\\
    \midrule
    \rowcolor{LightCyan}
    Airplane& Bicycle& Bird& Boat& Bottle& Bus& Car& Cat& Chair& Cow\\ 
    \rowcolor{LightCyan}
    Dining table& Dog& Horse& Motorcycle& Person& Potted plant& Sheep& Couch& Train& Tv\\ 
    \rowcolor{LightYellow}
    Truck& Traffic light& Fire hydrant& Stop sign& Parking meter& Bench& Elephant& Bear& Zebra& Giraffe\\ 
    \rowcolor{LightYellow}
    Backpack& Umbrella& Handbag& Tie& Suitcase& Microwave& Oven& Toaster& Sink& Refrigerator\\ 
    \rowcolor{LightYellow}
    Frisbee& Skis& Snowboard& Sports ball& Kite& Baseball bat& Baseball glove& Skateboard& Surfboard& Tennis racket\\ 
    \rowcolor{LightYellow}
    Banana& Apple& Sandwich& Orange& Broccoli& Carrot& Hot dog& Pizza& Donut& Cake\\ 
    \rowcolor{LightYellow}
    Bed& Toilet& Laptop& Mouse& Remote& Keyboard& Cell phone& Book& Clock& Vase\\ 
    \rowcolor{LightYellow}
    Scissors& Teddy bear& Hair drier& Toothbrush& Wine glass& Cup& Fork& Knife& Spoon& Bowl\\ 
    \bottomrule
    \vspace{0.1em}
  \end{tabular}
  }
  \resizebox{1 \linewidth}{!}{
    \begin{tabular}{cccccccccc}
    \toprule
    \rowcolor{LightRed}
    \multicolumn{10}{c}{Task-2}\\
    \midrule
    \rowcolor{LightCyan}
    Airplane& Bicycle& Bird& Boat& Bottle& Bus& Car& Cat& Chair& Cow\\ 
    \rowcolor{LightCyan}
    Dining table& Dog& Horse& Motorcycle& Person& Potted plant& Sheep& Couch& Train& Tv\\ 
    \rowcolor{LightCyan}
    Truck& Traffic light& Fire hydrant& Stop sign& Parking meter& Bench& Elephant& Bear& Zebra& Giraffe\\ 
    \rowcolor{LightCyan}
    Backpack& Umbrella& Handbag& Tie& Suitcase& Microwave& Oven& Toaster& Sink& Refrigerator\\ 
    \rowcolor{LightYellow}
    Frisbee& Skis& Snowboard& Sports ball& Kite& Baseball bat& Baseball glove& Skateboard& Surfboard& Tennis racket\\ 
    \rowcolor{LightYellow}
    Banana& Apple& Sandwich& Orange& Broccoli& Carrot& Hot dog& Pizza& Donut& Cake\\ 
    \rowcolor{LightYellow}
    Bed& Toilet& Laptop& Mouse& Remote& Keyboard& Cell phone& Book& Clock& Vase\\ 
    \rowcolor{LightYellow}
    Scissors& Teddy bear& Hair drier& Toothbrush& Wine glass& Cup& Fork& Knife& Spoon& Bowl\\ 
    \bottomrule
    \vspace{0.1em}
  \end{tabular}
  }
  \resizebox{1 \linewidth}{!}{
    \begin{tabular}{cccccccccc}
    \toprule
    \rowcolor{LightRed}
    \multicolumn{10}{c}{Task-3}\\
    \midrule
    \rowcolor{LightCyan}
    Airplane& Bicycle& Bird& Boat& Bottle& Bus& Car& Cat& Chair& Cow\\ 
    \rowcolor{LightCyan}
    Dining table& Dog& Horse& Motorcycle& Person& Potted plant& Sheep& Couch& Train& Tv\\ 
    \rowcolor{LightCyan}
    Truck& Traffic light& Fire hydrant& Stop sign& Parking meter& Bench& Elephant& Bear& Zebra& Giraffe\\ 
    \rowcolor{LightCyan}
    Backpack& Umbrella& Handbag& Tie& Suitcase& Microwave& Oven& Toaster& Sink& Refrigerator\\ 
    \rowcolor{LightCyan}
    Frisbee& Skis& Snowboard& Sports ball& Kite& Baseball bat& Baseball glove& Skateboard& Surfboard& Tennis racket\\ 
    \rowcolor{LightCyan}
    Banana& Apple& Sandwich& Orange& Broccoli& Carrot& Hot dog& Pizza& Donut& Cake\\ 
    \rowcolor{LightYellow}
    Bed& Toilet& Laptop& Mouse& Remote& Keyboard& Cell phone& Book& Clock& Vase\\ 
    \rowcolor{LightYellow}
    Scissors& Teddy bear& Hair drier& Toothbrush& Wine glass& Cup& Fork& Knife& Spoon& Bowl\\ 
    \bottomrule
  \end{tabular}
  }
  \caption{Semantic splits for Task-1, Task-2 and Task-3. Known classes are highlighted in blue. Unknown classes are highlighted in yellow.}
  \label{tab:t1}
\end{table*}

\section{Qualitative Analysis}
\label{sec:Qualitative}
\subsection{Open-Set Detection and Discovery Results}
\begin{figure*}[t]
  \centering
   \includegraphics[width=1.\linewidth,page=1]{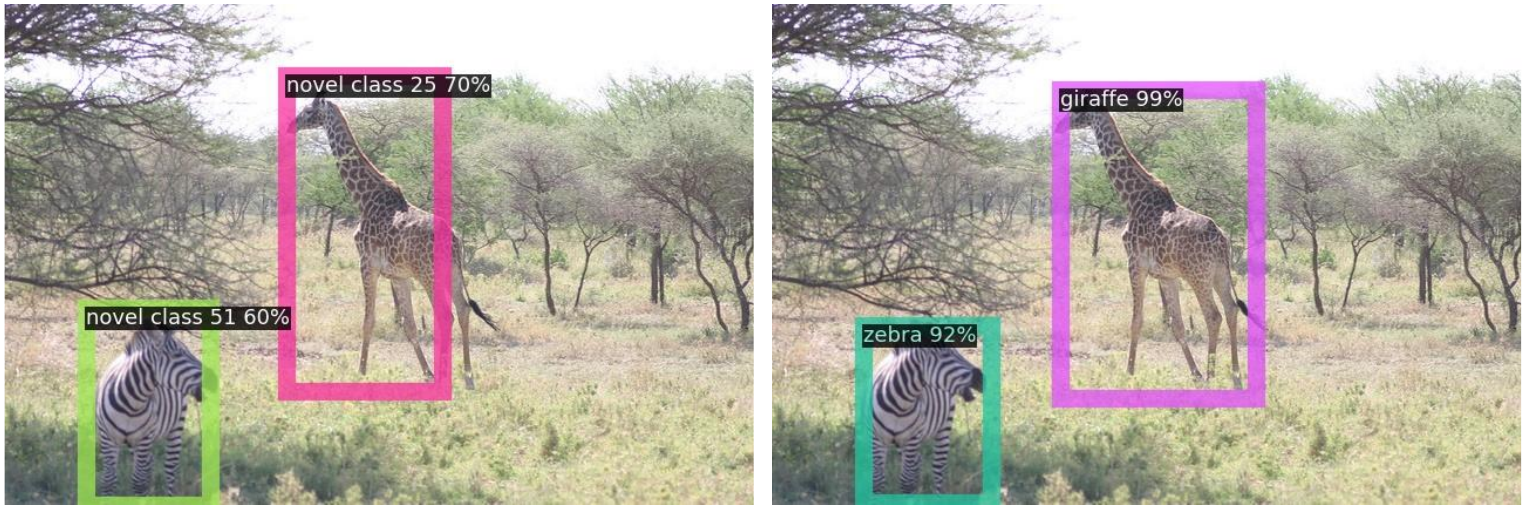}
   \caption{OSODD predictions in Task-1~(Left) and Task-2~(Right). In Task-1, the model successfully localises the unknown objects and recognise them as two different categories. In Task-2, the wild animal classes including \textit{zebra} and \textit{giraffe} are introduced to model, it correctly classifies the objects into their corresponding classes.}
   \label{fig:fig_qua_1}
\end{figure*}

\begin{figure*}[t]
  \centering
  \resizebox{0.77 \linewidth}{!}{
   \includegraphics[width=1.0\linewidth,page=5]{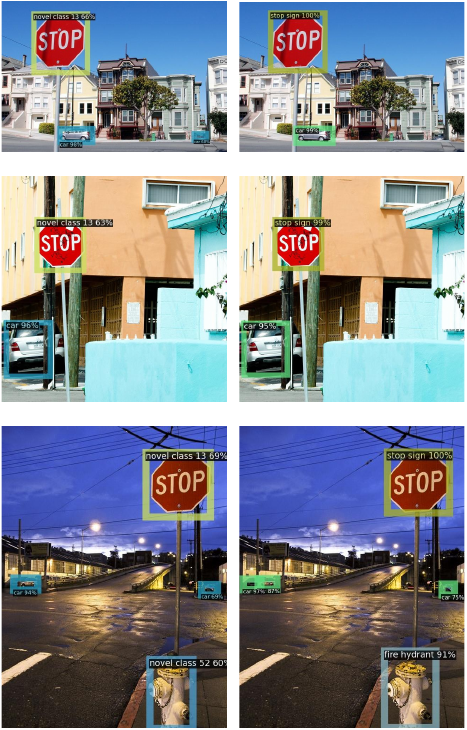}
   }
   \caption{The left column shows the results in Task-1 where only 20 classes are available. The right column shows the results in Task-2 where 20 new classes, like \textit{stop sign} and \textit{fire hydrant}, have been introduced to the model.}
   \label{fig:fig_qua_2}
\end{figure*}

\begin{figure*}[t]
  \centering
  \resizebox{0.77 \linewidth}{!}{
   \includegraphics[width=1.0\linewidth,page=8]{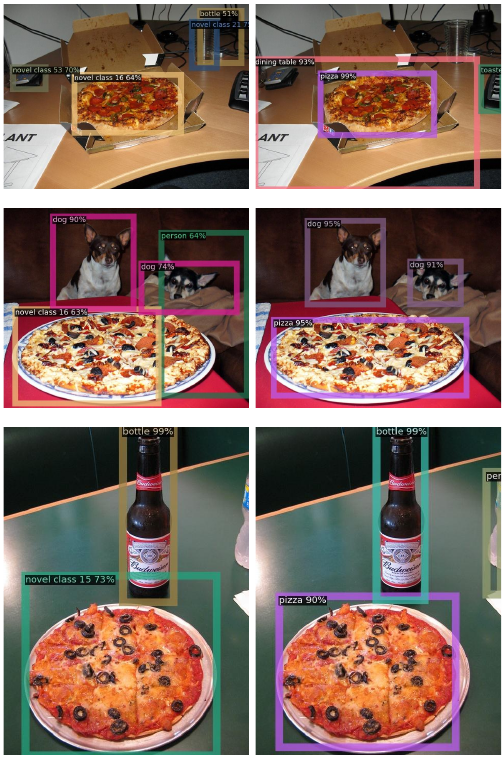}
  }
   \caption{The left column shows the predictions in Task-1 where the \textit{piazzas} are clustering into two novel categories. The right column shows the predictions in Task-3 where all the \textit{piazza} objects are correctly classified.}
   \label{fig:fig_qua_3}
\end{figure*}

\begin{figure*}[t]
  \centering
  \resizebox{0.65 \linewidth}{!}{
   \includegraphics[width=1.0\linewidth,page=6]{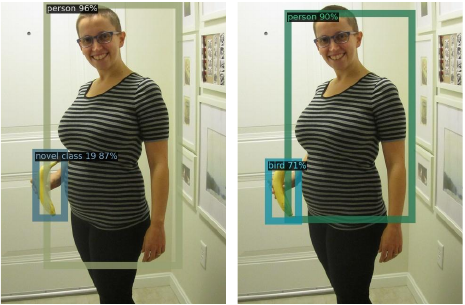}
  }
   \caption{The left column shows the predictions in Task-1. The banana has not been introduced, the model has correctly predicted the novel categories. The right column shows the predictions in Task-2 where the food classes are not available for the task. The model should predict the banana into one of the novel categories, but it incorrectly classifies the unknown object into one of the known classes.}
   \label{fig:fig_qua_4}
\end{figure*}

\begin{figure*}[t]
  \centering
  \resizebox{0.65 \linewidth}{!}{
   \includegraphics[width=1.0\linewidth,page=7]{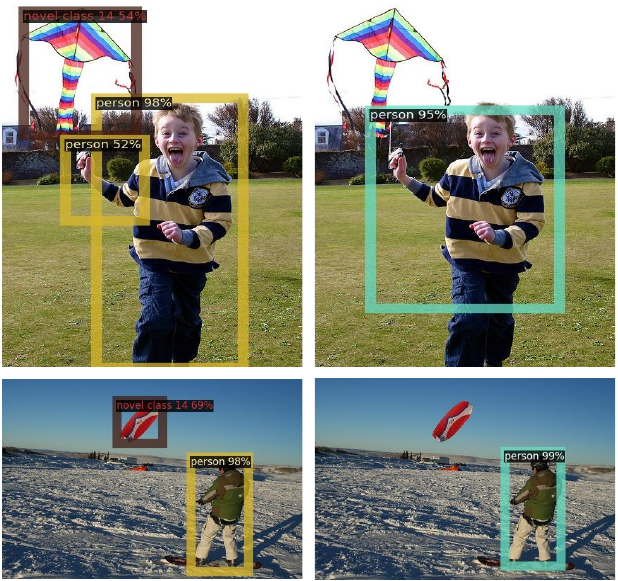}
  }
  \caption{A failure case in open-set learning. The model successfully discovers the novel category for the \textit{kite} class in Task-1~(Left). However, after more semantic classes of labels are provided in the following task, Task-2, the model fails to localise the \textit{kites} in the image.}
   \label{fig:fig_qua_5}
\end{figure*}

In \cref{fig:fig_qua_1}, we visualise the OSODD predictions under two tasks, Task-1 and Task-2. The left figure shows the prediction in Task-1, where the \textit{zebra} and \textit{giraffe} class are not introduced. The model successfully distinguishes two unknown animals. The right figure shows the prediction in Task-2 where the annotations of \textit{zebra} and \textit{giraffe} are made available. 
More results are shown in~\cref{fig:fig_qua_2}.
We have also encountered failure cases, as shown in~\cref{fig:fig_qua_3}, where the model incorrectly assigns the \textit{piazzas} to two novel categories. Additionally, in the second row, there is a false detection on \textit{person} class.
However, after the \textit{piazza} class get introduced in Task-3, the model makes the correct predictions on all \textit{piazza} instances and eliminates the ambiguity of the novel categories. Two different failure cases are shown in \cref{fig:fig_qua_4} and \cref{fig:fig_qua_5}. In the first case~(See \cref{fig:fig_qua_4}), the detector incorrectly classifies the unknown objects as the known classes. In the second case~(See \cref{fig:fig_qua_5}), the detector correctly finds the novel category for the unknown objects when the labels are not available, but it does not detect the objects when the actual class is introduced.
This suggests there is still a large space to be improved.

\subsection{Object Category Discovery Results}
\begin{figure*}[t]
  \centering
   \includegraphics[width=1.0\linewidth,page=1]{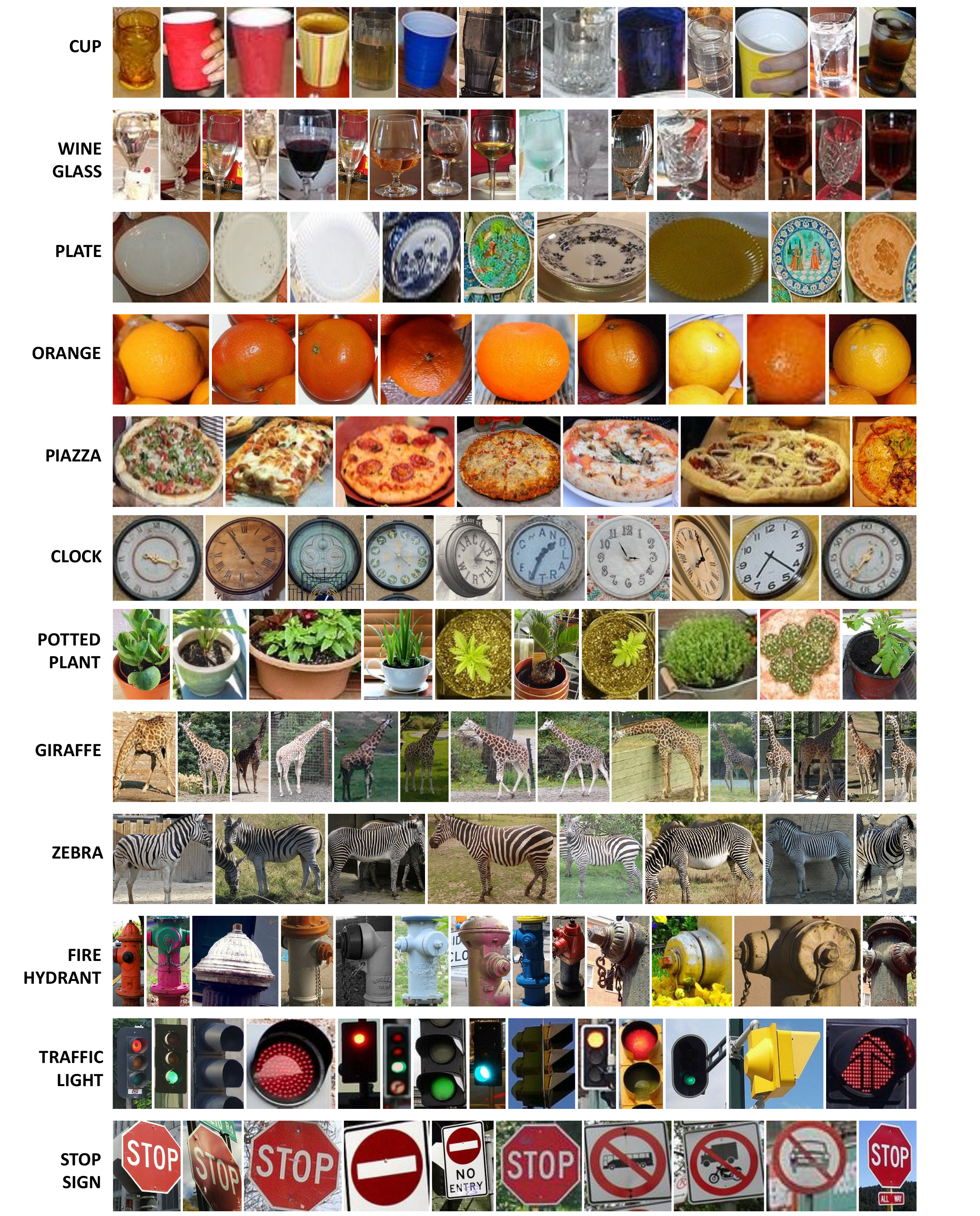}
   \caption{Some discovered results from object category discovery.}
   \label{fig:fig_qua}
\end{figure*}

In~\cref{fig:fig_qua}, we visualise some discovered object clusters from the first task, Task-1.
We assume that 20 classes are known and the rest 60 classes are unknown (See `Task-1' in~\cref{tab:t1}). Most clusters can be quantitatively evaluated by the ground-truth labels in the validation step. Some objects or categories of interest are not annotated by human in the original dataset (\eg plate). Surprisingly, our model can identify those un-annotated objects and cluster them to find new novel categories (See `PLATE' in \cref{fig:fig_qua}). 
It is noticed that some objects from the known class have been falsely predicted as unknown and clustered into novel categories (\eg potted plant).



\end{appendices}

\end{document}